\documentclass[letterpaper, 10 pt, conference]{ieeeconf}  
\usepackage[spanish]{babel}
\usepackage[utf8]{inputenc}
\usepackage{listings}
\usepackage{caption}
\usepackage{DejaVuSans}
\usepackage{DejaVuSansMono}
\usepackage{amssymb}
\usepackage{amsmath}
\usepackage{lipsum}
\usepackage{subcaption}
\usepackage{graphicx}

\usepackage{alltt}

\usepackage{fancyvrb}

\usepackage[table,xcdraw]{xcolor}
\usepackage{multirow}

\usepackage[T1]{fontenc}
\usepackage{fancyvrb}
\usepackage{listings}
\usepackage[scaled=.55]{beramono}    
\fvset{baselinestretch=0.5}
\DeclareCaptionFormat{listing}{\rule{\dimexpr0.9\columnwidth+17pt\relax}{0.35pt}\par\vskip1pt#1#2#3}
\newsavebox{\FVerbBox}
\usepackage[most]{tcolorbox}
\tcbset{
    frame code={}
    center title,
    left=0pt,
    right=0pt,
    top=0pt,
    bottom=0pt,
    colback=gray!10,
    colframe=white,
    width=0.45\textwidth,
    enlarge left by=0mm,
    boxsep=5pt,
    arc=0pt,outer arc=0pt,
    }

\IEEEoverridecommandlockouts                              

\title{\LARGE \bf
Time Synchronization in modular collaborative robots
}
\author{
\textbf{Carlos San Vicente Gutiérrez}, 
\textbf{Lander Usategui San Juan}, 
\textbf{Irati Zamalloa Ugarte}, \\
\textbf{Iñigo Muguruza Goenaga},
\textbf{Laura Alzola Kirschgens} and 
\textbf{Víctor Mayoral Vilches}\\
Erle Robotics S.L.\\
Vitoria-Gasteiz, Alava, Spain\\
}

\usepackage{blindtext}
\usepackage{graphicx}
\usepackage{listings}
\usepackage{xcolor}

\PassOptionsToPackage{hyphens}{url}\usepackage{hyperref}
\hypersetup{
    colorlinks=true,
    linkcolor=blue,
    filecolor=magenta,      
    urlcolor=cyan,
    citecolor=blue,
}
\PassOptionsToPackage{hyphens}{url}\usepackage{hyperref}

\usepackage[normalem]{ulem}

\begin{document}
\maketitle

\begin{abstract}

A new generation of robot systems which are modular, flexible and safe for human-robot interaction are needed. Existing cobots seem to meet only the later and require a modular approach to improve their reconfigurability and interoperability. We propose a new sub-class of cobots named \emph{M-cobots} which tackle these problems. In particular, we discuss the relevance of synchronization for these systems, analyze it and demonstrate how with a properly configured \emph{M-cobot}, we are able to obtain a) distributed sub-microsecond clock synchronization accuracy among modules, b) timestamping accuracy of ROS 2.0 messages under 100 microseconds and c) millisecond-level end-to-end communication latencies, even when disturbed with networking overloads of up to 90\% of the network capacity.


\end{abstract}

\section{Introduction}
\label{introduction}


With the growing popularity of robots, we are starting to observe how these machines are entering new areas. The degree of automation in a variety of applications is rapidly increasing. The food industry is one of the latest hypes in robotics and current technology is showing limitations to cope with the demands that arise when \emph{preparing a salad}. Moving beyond new robot applications and looking back at traditional industrial automation, technical hurdles are starting to be identified as critical for further growth. Geenen \cite{modflex} indicates that previous work conducted at Fraunhofer \cite{muller2013wandlungsfahiges} showed that full automation of production cycles is often inefficient due to the challenges presented when mixing high numbers of product varieties combined in small batch quantities. He remarks that achieving efficiency in this landscape requires the development of flexible and adaptable production systems that can be applied to a variety of tasks \cite{corves2012reconfigurable} and proposes a new generation of robot systems which are modular, flexible and safe for Human-Robot-Collaboration (HRC).\\

Collaborative robots (commonly referred as \emph{cobots}), combine the benefits of human intelligence and skills with the advantage of sophisticated robotic technical systems. This generation of robots focus on establishing a joint workspace between humans and machines. Cobots  have demonstrated their advantages in many areas.  For example, Gambao et al. highlight this aspect for material handling tasks \cite{gambao2012new} however, according to the authors, in order to improve the reconfigurability and flexibility of existing robots, the modular approach is the best solution. Current available-in-the-market collaborative robots lack of modularity and have followed a similar approach than that of traditional robots enforcing vendor lock-in through a variety of techniques. Beyond small agrupations of manufacturers around an individual cobot vendor, there is no real attempt to ease the integration process with peripherals such as end-effectors or sensors in a vendor-agnostic manner.\\

Inspired by prior art \cite{gambao2012new}, we present a new subclass of collaborative robots that include native modularity and reconfigurability capabilities to tackle new market demands.  We call this new class \emph{M-cobots}; modular collaborative robots. \emph{M-cobots} aim to tackle several of the relevant problems that exist in current cobot solutions such as the lack of interoperability among vendors, the lack of flexibility, extensibility or even synchronization with third party modules. In this particular paper, we discuss the problem of synchronization, relevant for time-critical robotic tasks and with direct implications on communication latencies. \\

Traditional approaches for synchronization involve the use of certain vendor-specific fieldbuses and a variety of components using different synchronization primitives. This, typically leads to complex and hard to maintain hybrid synchronization architectures. For \emph{M-cobots}, we propose an integrated synchronization approach based on the IEEE 1588 Precision Time Protocol (PTP) standard. We demonstrate how with this approach, we manage to obtain robust millisecond-level end-to-end communication latencies. We show our setup and provide experimental results about the timestamping accuracy obtained with ROS 2.0. In particular, we show how our distributed and modular setup is able to deliver sub-microsecond clock synchronization accuracy. We finalize by challenging our setup with a variety of actions and showing how appropriate configurations can mitigate the impact to the overall communication latency.\\

The content is structured as follows: section \ref{background} will introduce an overview of different synchronization approaches in distributed robotic systems. Section \ref{setup_results} discusses the experimental results obtained while evaluating the presented hypotheses. Finally, section \ref{conclusions} presents our conclusions and future work.\\

\section{Background} 
\label{background}



\subsection{Why do we need synchronization in modular robotics?} 
\label{}

There is abundant literature on the importance of synchronization in robotics applications and how different methods are applied in order to achieve better performance. A modular robotic system is, in essence, a distributed robotic system which requires synchronization between components in order to coordinate operations for the sensors and the actuators. A common time reference is required for a variety of reasons. For example, often, data logging requires precise timestamping. In high speed and accurate motion control systems, synchronization between different actuators is critical. In the case of multiple sensor data fusion, inaccurate synchronization usually leads to inaccurate data inference. Furthermore and as it will be seen through our experiments, synchronization is also critical for decreasing overall response times in distributed systems.\\


A common case in robotics where synchronization is needed are robot-laser scanner systems. Voges et al. \cite{Voges_Wieghardt_Wagner_2018} explain the significance of a proper synchronization between a laser scanner and an actuator within a motor-actuated 3D laser scanner. In this case, the authors propose a passive method to estimate the sensor timestamp offset. In \cite{timestamp}, a similar method is applied for a multi-sensor system consisting in an actuated laser scanner, its motor and a camera for a Simultaneous Localization and Mapping (SLAM) application. As the author explains, if no synchronization is present, the offset for the corresponding encoder values for each set of scan points can lead to a large distortion in the resulting point cloud that is constructed by the SLAM technique.\\

In \cite{6393361}, it is explained how the synchronization between a laser scanner and a robot is critical for laser welding. An incorrect synchronization would lead to geometric error between the desired pattern and the welded pattern. In this case, the proposed method consisted in characterizing experimentally and compensating the delay between measurements and robot operations. Graaf et al. \cite{welding_tracking} tackles a similar problem for real-time seam-tracking in robotic laser welding using a UDP communication to synchronize the laser data acquisition with the robot pose.\\ 

Zaman et al. \cite{zaman} introduces a method for time synchronization of odometry and vision data in a mobile robot. The author explains how in systems without synchronization techniques, the combination of image data with odometry usually requires speed limitations in order to reduce measurements errors. The proposed method use motion events, detectable by the odometry and image data.\\

In \cite{humanoid}, a humanoid robot is built as a distributed system using Ethernet and a communication middleware (Corba) for the communications. This work addresses the importance of synchronization and uses the IEEE 1588 Precision Time Protocol (PTP) to synchronize the clock of all the robot components in the network. Particularly, within the example discussed, the nodes need to be synchronized in order to get data from the sensors and rotate the motors at the right time.\\

In \cite{multi_sensor_nico} different synchronization strategies for multi-sensor data fusion for advanced driver assistant systems (ADAS) applications are discussed. The aim of this work is to analyze in a formal way the impact of the sensor fusion process in the system latency. One of the final conclusions is that having the sensors synchronized helps to reduce the fusion process latency greatly in scenarios with very different sensor measuring frequencies. \\

In \cite{ros_thesis}, the use of real-time sensor data for robot control is analyzed. In this work, the author breaks down the delays involved in a sensor based control with a Universal Robot (UR) arm controlled using the Robot Operating System (ROS) \cite{quigley2009ros}. For the characterization, the worst case synchronization delays are taken into account. These delays are as high as the period of the control cycle. In order to compensate for sampling effects due to the lack of synchronization between the sensors and the control cycle, an oversampling strategy is adopted. With this strategy the sensors values are updated with a sample frequency higher than that of the robot control cycle.\\ 

As we have seen, synchronization is required in a wide range of robotic applications. Depending on the use case, it is common to find different solutions; some of them are based on clock offset estimation and compensation, calibration, trigger signals and time synchronization standards, among others.\\

\subsection{Synchronization in ROS}
\label{ros_sync}

In ROS based systems, the common approach for synchronization is to use the system clock as a time source and to synchronize the clocks of different components with network synchronization tools such as the Network Time Protocol (NTP), as explained in ROS clock documentation \cite{ros_clock}. In addition, ROS provides several topic synchronization methods as part of the Transform Library \cite{tf_Foote}. \texttt{tf} or the newer \texttt{tf2} libraries have methods such as \texttt{TimeSynchronizer} and \texttt{ApproximateTimeSynchronizer}, which allows to synchronize incoming messages based on the timestamps in the message headers. ROS 2.0 Clock and Transform Library implementations are currently under development \cite{ros2_clock} but it is expected to work in a similar way as in ROS.\\



One of the problems when building a ROS based system is that synchronizing the system clocks is not always feasible and in other cases the method used is not accurate enough. For this reason, it is common to find different approaches to achieve synchronization between components. Olson et al. \cite{olson} give an overview of some of these methods and propose a passive synchronization algorithm. This method aims to reduce the synchronization error in robot sensors when the sensors do not provide a synchronization method. In this work, one of the main problems in robotic system construction is addressed; heterogeneous sensors which lack of synchronization standards. For example, in many cases, synchronization methods such as NTP are not applicable because sensors are connected trough serial interfaces. This problem is also addressed in the 2017 Roscon talk \emph{Building a Computer Vision Research Vehicle with ROS} \cite{roscon_daimler} where Fregin explained the method used to solve the synchronization problem between four cameras for computer vision system for an ADAS (Advanced driver-assistance systems) system. Because the cameras lack of a common reference time, they needed to implement a hybrid solution based on the PTP and a trigger signal approach. This exemplifies the complexity of integrating non-synchronized components in a ROS distributed system.\\

As discussed, synchronization can be time consuming and the implemented solutions tend to be application specific. The key problem is that many robot components do not provide standard synchronization methods or they do not provide any synchronization method at all. To the best of our knowledge, the simplest approach for roboticists would be to use synchronous ROS native components which can be easily synchronized using enabled methods such as NTP or PTP. This way, ROS applications will simply rely on a correct system clock synchronization regardless of the method used to synchronize the clocks.\\

\subsection{Precision Time Protocol (PTP)}

NTP is a well established clock synchronization method which allows synchronization over unreliable networks such as the Internet. On the other hand, when higher accuracy is needed, clock synchronization protocols such as Global Positioning System (GPS) are typically used. However, to overcome some limitations of these protocols the IEEE 1588 Precision Time Protocol (PTP) was created. PTP provides higher acuracy than NTP because it makes use of hardware timestamping, which provides sub-microsecond accuracy \cite{ntp_vs_ptp}. Also, unlike GPS, PTP does not require to have access to satellite signals, which makes it suitable for indoors applications and at lower cost.\\

%

In order to simplify the construction of synchronous robots powered by ROS, we make use of the H-ROS \cite{HROS} software and hardware infrastructure and include the IEEE 1588 Precision Time Protocol (PTP) at each module. In addition, the H-ROS modules make use of the HRIM \cite{zamalloa2018information} model which allows modules to have consistent messages and interoperate seamlessly. The PTP standard included on each module allows sub-microsecond synchronization. This provides a common global clock for all the H-ROS modules allowing, for example, to timestamp all the messages in a coherent time reference.\\


\subsection{Time synchronization and TSN}
\label{tsn_ptp}

As explained in previous work \cite{DBLP:journals/corr/abs-1804-07643}, the Time Sensitive Networking (TSN) standards aim to make Ethernet more deterministic. As part of these standards, the IEEE 802.1 TSN task group has defined a formal profile of IEEE 1588, in order to synchronize time in standard Ethernet. The IEEE 802.1AS-2011 \cite{802.1AS} standard, also known as gPTP (Generalized Precision Time Protocol), defines a specific profile of IEEE 1588-2008 with additional specifications which improve timing accuracy and reliability. Following the completion of 802.1AS-2011, the IEEE 802.1 TSN task group began work on a revision of gPTP in order to expand its use for industrial time-sensitive applications. For example 802.1AS-REV will support multiple simultaneous master clocks to allow for quick recovery of time synchronization in case of failure. The standard also aims to provide plug and play synchronizations, an interesting feature for modular robotic components.\\

Time synchronization provided by the 802.1AS standard is also the base for other standards such as the time aware traffic scheduling, defined in 802.1Qbv \cite{802.1Qbv}. This standard enables time-triggered communications at the egress ports of Ethernet devices by defining a periodic schedule.\\

\section{Experimental setup and results}
\label{setup_results}

\begin{figure}[h!]
\centering
 \includegraphics[width=0.4\textwidth]{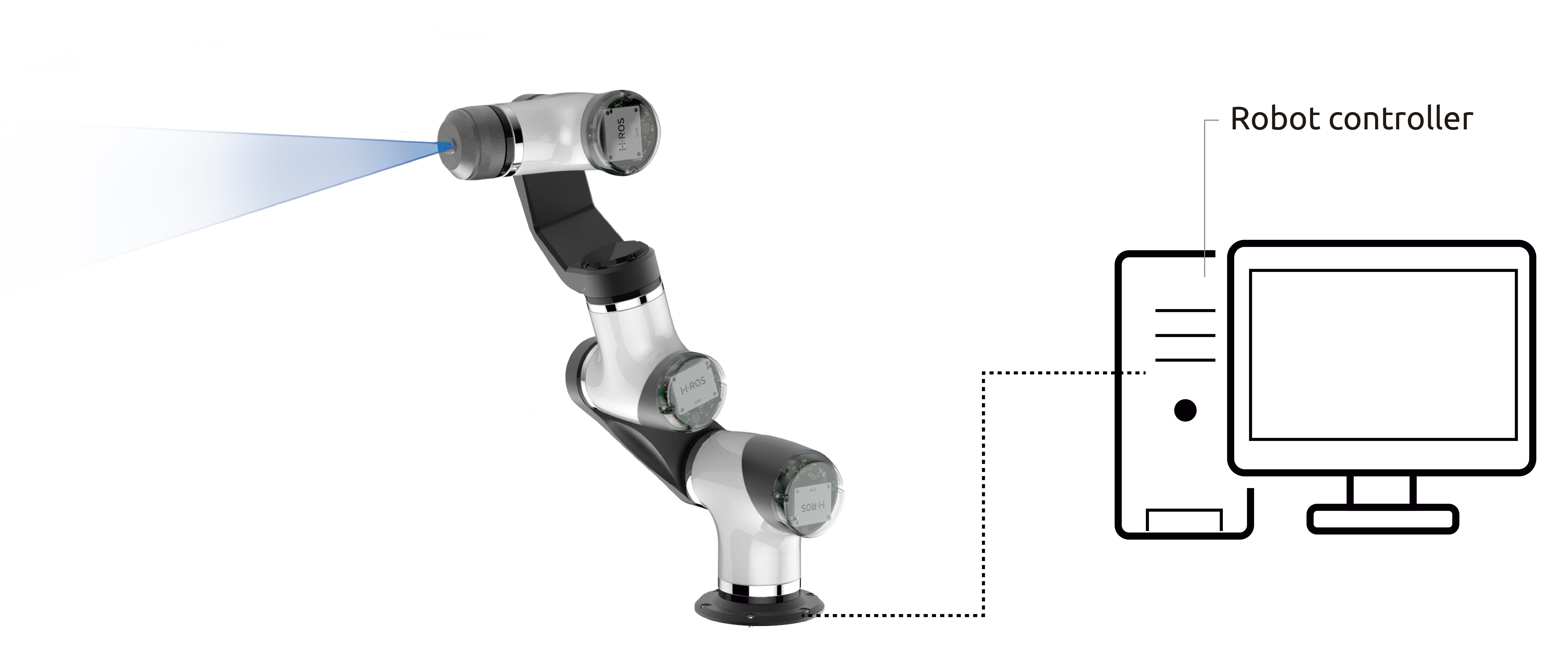}
\caption{\footnotesize Experimental setup with a 6 Degrees-of-Freedom (DoF) modular articulated arm and a range finder}
\label{setup}
\end{figure}

The aim of the experiments is to characterize the synchronization and communications delays of a ROS 2.0 distributed modular robotic system. The robotic system used for the experiments is compound by a 6 Degrees-of-Freedom (DoF) modular articulated arm with a range finder. The modular robotic arm has been built using the H-ROS\cite{HROS} technology, which simplifies the process of building, configuring and re-configuring robots. For the actuators of the arm, we used three two-joint servo motors modules. Each module includes an embedded device with a ROS 2.0 native system able to publish the status of the motors trough HRIM\cite{hrim} rotary servo messages. All the motor modules and the range finder are connected trough Ethernet cables in daisy-chaining, as shown in figure \ref{setup}.

The robot controller is implemented in a PC with the following characteristics:
\begin{itemize}
    \item Processor: Intel(R) Core(TM) i7-8700K CPU @ 3.70GHz (6 cores).
    \item OS: Ubuntu 16.04 (xenial).
    \item ROS 2 version: Bouncy Bolson.
    \item Kernel version: 4.9.30.
    \item PREEMPT-RT patch: RT21.
    \item Link capacity: 100/1000 Mbps, Full-Duplex.
    \item Network interface card: Intel i210.
\end{itemize}

In the other hand, the main characteristics of the embedded device are:
\begin{itemize}
    \item Processor: ARMv7 Processor (2 cores).
    \item OS: Linux-based
    \item ROS 2.0 version: Bouncy Bolson.
    \item Kernel version: 4.9.30.
    \item PREEMPT-RT patch: RT21.
    \item Link capacity: 100/1000 Mbps, Full-Duplex.
\end{itemize}

\subsection{Timing measurements}

\begin{figure}[h!]
\centering
 \includegraphics[width=0.45\textwidth]{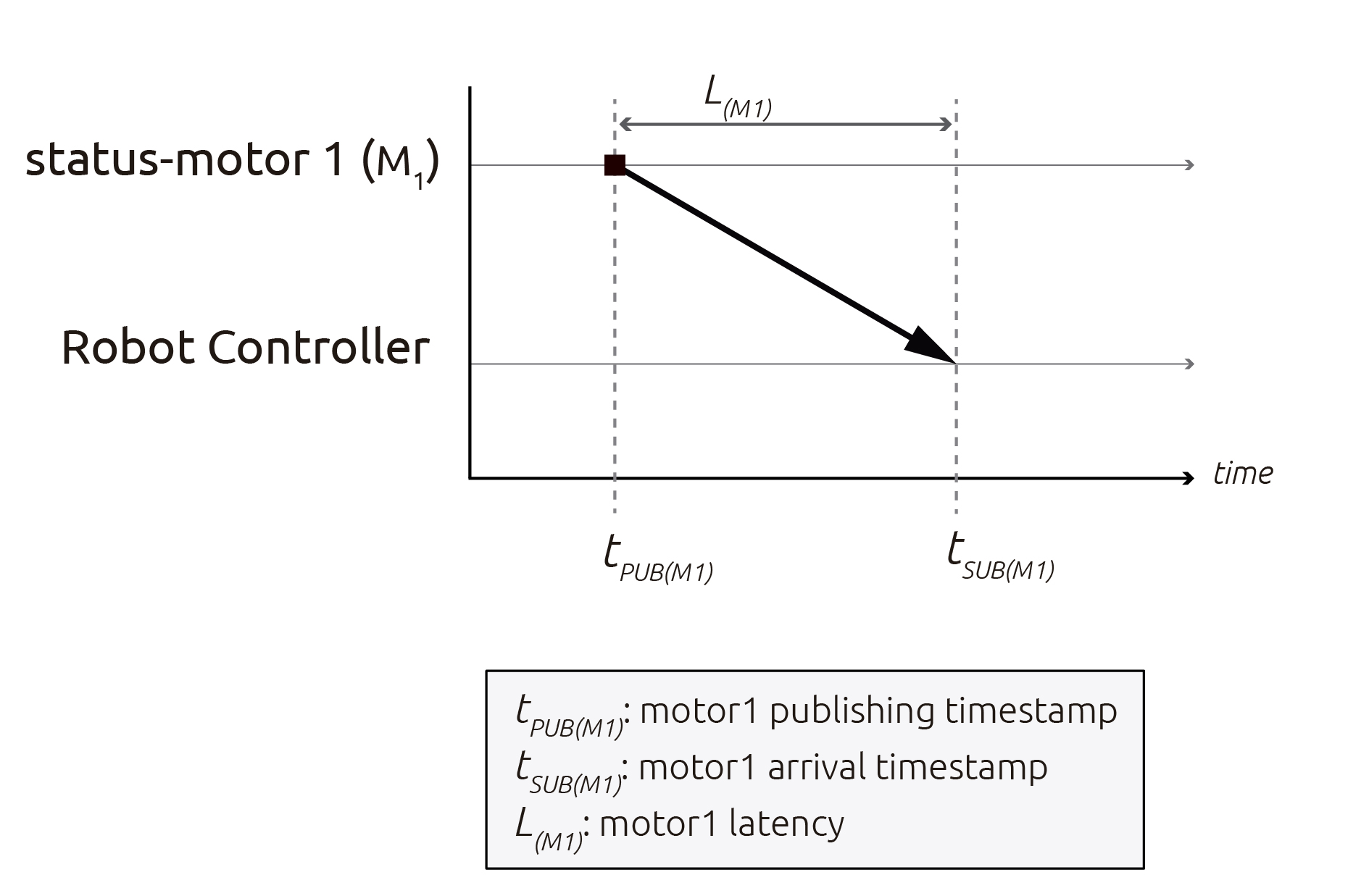}
\caption{\footnotesize Timing measurements using one-way timestamping.}
\label{One_way_timestamping}
\end{figure}

In our experiments, we analyze the timing for the communications from the robot and sensors to the robot controller. For simplicity, we are not analyzing the communications from the robot controller to the actuators; however, the results should be equivalent. In order to characterize the communications we use a one-way timestamping strategy, as shown in figure \ref{One_way_timestamping}. Each message is timestamped before calling the ROS 2.0 publish function ($t_{_{PUB}}$). This timestamp is written in a specific variable of the \texttt{std\_msgs} called header. In the robot controller, the arrival timestamp ($t_{_{SUB}}$) is measured at the beginning of the topic subscription callback. In the callback, both timestamps are stored in a buffer to log them in a file and calculate the statistics.\\




\begin{figure}[h!]
\centering
 \includegraphics[width=0.5\textwidth]{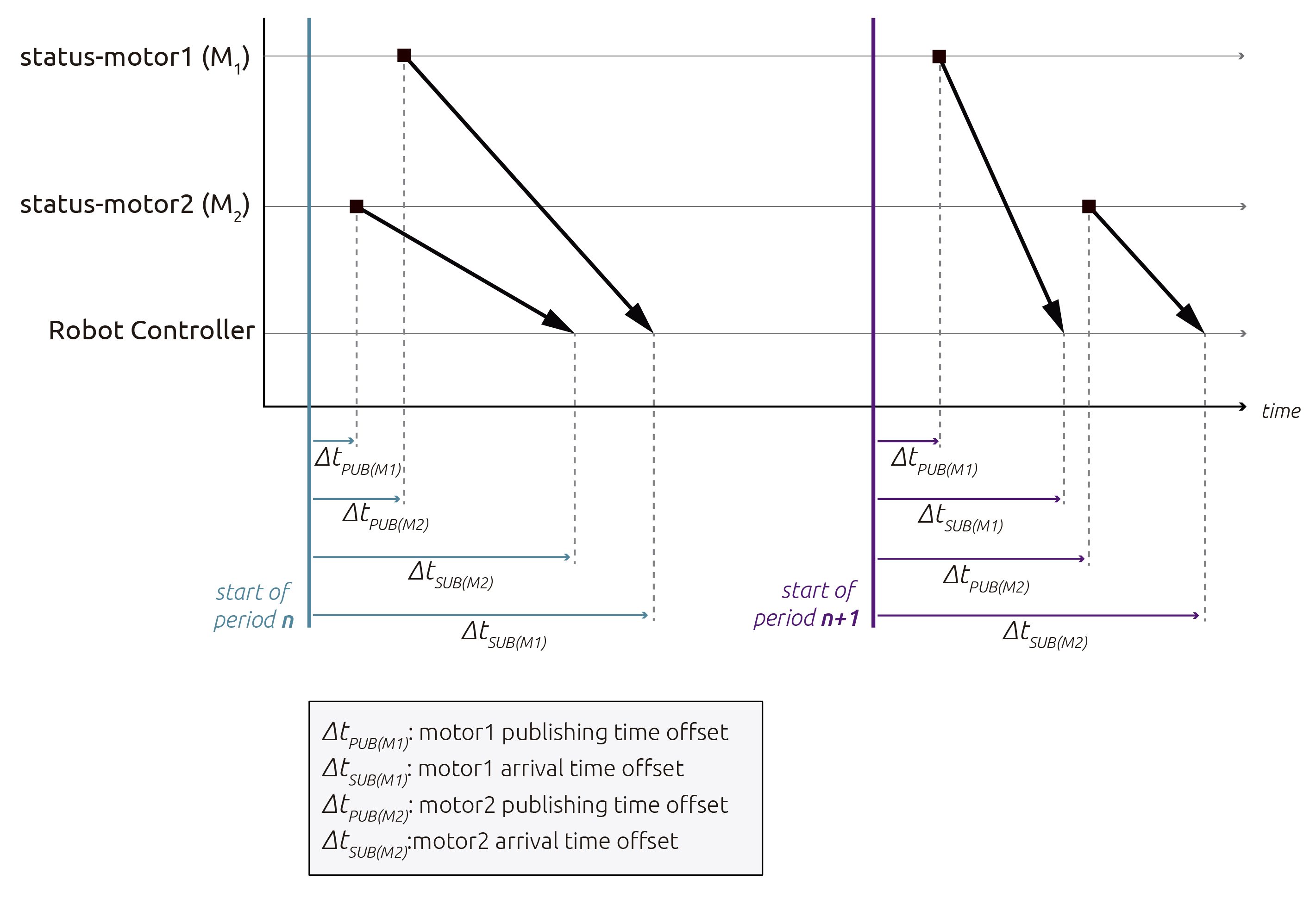}
\caption{\footnotesize Timing measurements for two consecutive periods using system clock timestamps for two servo motors.}
\label{matched_timestamps}
\end{figure}

In order to compare samples from different periods we use the publishing period rate as a reference. The publishing and arrival timestamps are analyzed observing the offsets with respect the start of each period. We define the start of the period by convention\footnote{ We start all the publishing periods when the POSIX time reminder after division by the period (modulo operation) is equal zero. For example for a 100 millisecond rate the start of each period would follow the sequence \texttt{1535967970800000000}, \texttt{1535967970900000000}, \texttt{1535967971000000000}, etc.}. In figure \ref{matched_timestamps} we show an example of two motors measurements for two consecutive periods. By observing the publishing time offsets ($\Delta t_{_{PUB}}$) we can analyze the jitter and synchronization of the publishers. In the same way for the subscriptios, with the the arrival time offsets ($\Delta t_{_{SUB}}$) we can analyze the synchronization between topics. When the system clocks have been synchronized, the publishing and arrival timestamps can be compared and the transport delay calculated ($L$).\\



\subsection{Experimental results}

\subsubsection{Unsynchronized publishers}
\label{case1}

As a first naive approach, we make the measurements without synchronizing the system clocks of the modules and using the default ROS Timers. We use a 100 millisecond period for all the publishers and run the experiment for 10 minutes. Latency cannot be measured in this case, but we can measure when messages arrival offsets with respect to the robot controller systems clock and the publishing offsets with respect to the systems clock of each component.\\

\begin{figure}[h!]
\centering
 \includegraphics[width=0.4\textwidth]{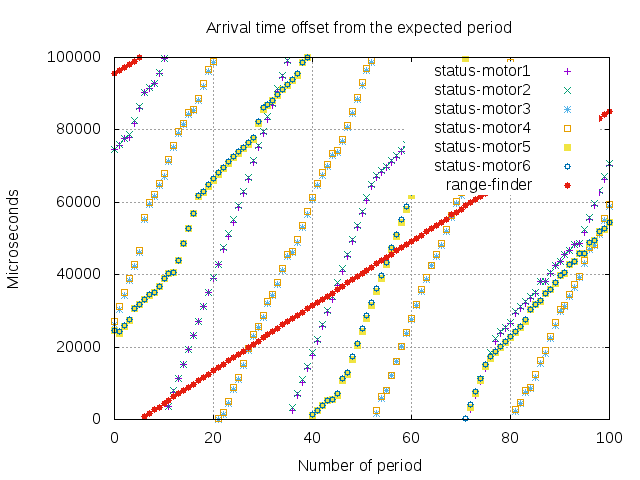}
\caption{\footnotesize Arrival time offset ($\Delta t_{_{SUB}}$) timeplot for a 100 millisecond period with ROS 2.0 timer and no synchronization.}
\label{time_rx_ros_timer}
\end{figure}

In the arrival timestamp ($\Delta t_{_{SUB}}$) timeplot (figure \ref{time_rx_ros_timer}) we observe that the offset and drift of the arrival times are considerable. For just 100 samples (10 seconds), we observe a drift of almost 100 milliseconds. However, this drift is not caused by clock misalignment but because we are using a relative sleep time for publishing.\\ 

\begin{figure}[h!]
\centering
 \includegraphics[width=0.4\textwidth]{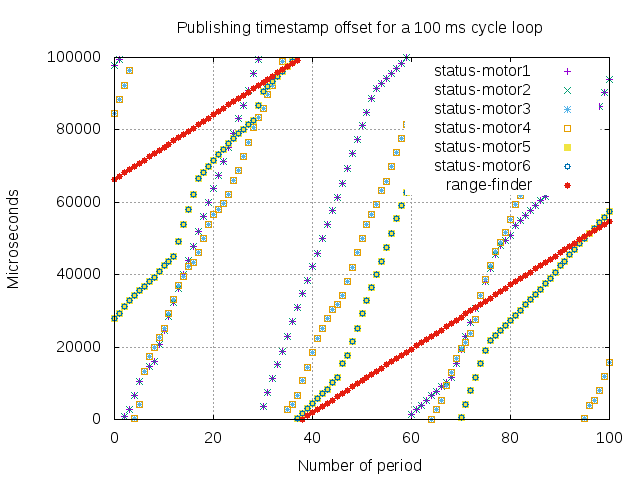}
\caption{\footnotesize Publishing time offset ($\Delta t_{_{PUB}}$) timeplot for a 100 millisecond period with ROS 2.0 timer and no synchronization.}
\label{time_tx_ros_timer}
\end{figure}

This can be observed in the publishing timestamps ($\Delta t_{_{PUB}}$) timeplot (figure \ref{time_tx_ros_timer}). We notice that the same drift observed in the reception plot. Relative timers are not recommended in order to implements a deterministic cyclic task because they cause a local drift over time, as explained in \cite{gallmeister1995posix} and \cite{clock_nanosleep} \footnote{The timing will drift the same amount of time it takes to call the publish tasks.}. To avoid this problem, an absolute timer must be used.\\

\subsubsection{Using an absolute timer}
\label{case2}
\begin{figure*}[h!]
  \begin{subfigure}[t]{.5\textwidth}
    \centering
    \includegraphics[width=0.8\linewidth]{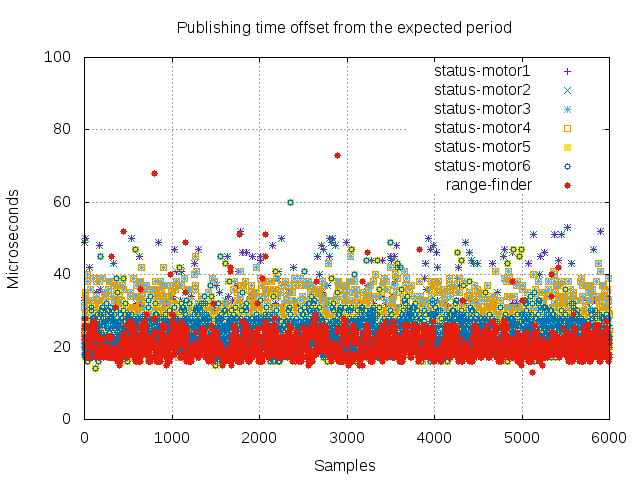}
    \caption{\footnotesize }
    \label{time_tx_abs_timer_be}
  \end{subfigure}
  \hfill
  \begin{subfigure}[t]{.5\textwidth}
    \centering
    \includegraphics[width=0.8\linewidth]{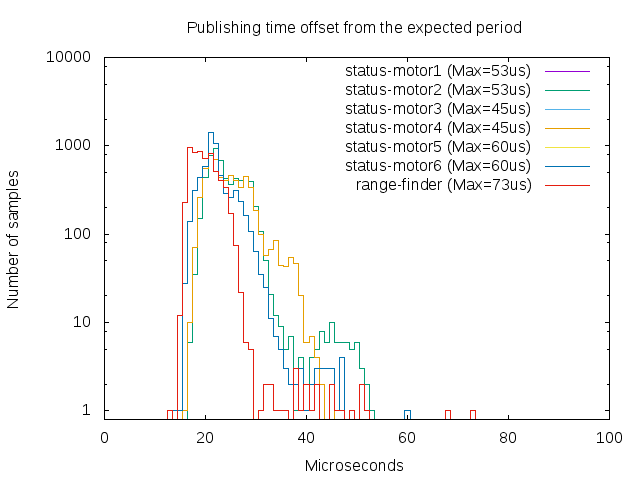}
    \caption{\footnotesize }
    \label{hist_tx_abs_timer_be}
  \end{subfigure}
  \caption{\footnotesize Measurements for a 100 millisecond period and 10 minutes duration period. a) Publishing time offset ($\Delta t_{_{PUB}}$) timeplot with an absolute timer and no synchronization b) Publishing time offset ($\Delta t_{_{PUB}}$) histogram with an absolute timer and no synchronization.}
\end{figure*}

As there is no absolute timer in the current ROS 2.0 timer implementation, we use the POSIX high-resolution sleep \texttt{clock\_nanosleep} enabling the \texttt{TIMER\_ABSTIME} flag. This will ensure repeatable publishing times with respect of the system clock of each component, as shown in figure \ref{time_tx_abs_timer_be} where the drift has been eliminated. The publishing timestamp presents very low jitter because we are using a real-time kernel. This is very important for applications that require accurate timestamping. Note that in this case we are writing the \emph{publication} timestamp in the message header for the measurements, but in general we would put the \emph{acquisition} timestamp in the message. \\

\begin{figure*}[h!]
  \begin{subfigure}[t]{.5\textwidth}
    \centering
    \includegraphics[width=0.8\linewidth]{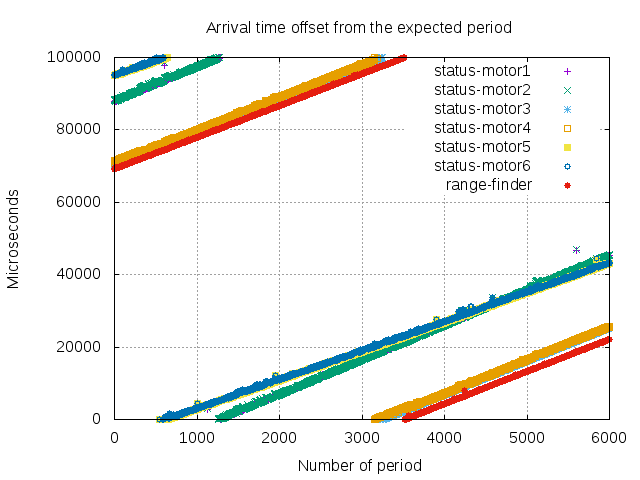}
    \caption{\footnotesize }
    \label{time_abs_timer_be}
  \end{subfigure}
  \hfill
  \begin{subfigure}[t]{.5\textwidth}
    \centering
    \includegraphics[width=0.8\linewidth]{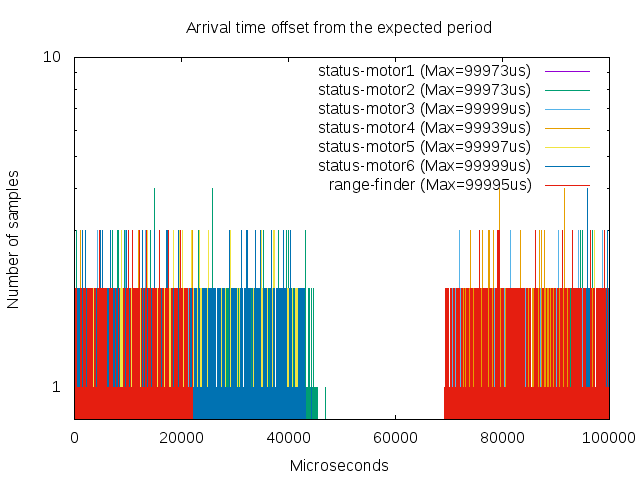}
    \caption{\footnotesize }
    \label{hist_rx_abs_timer_be}
  \end{subfigure}
  \caption{\footnotesize Measurements for a 100 millisecond period and 10 minutes duration period. a) Arrival time offset ($\Delta t_{_{SUB}}$) timeplot with an absolute timer and no synchronization b) Arrival time offset ($\Delta t_{_{SUB}}$) histogram with an absolute timer and no synchronization.}
\end{figure*}

In the arrival timestamp offset ($\Delta t_{_{SUB}}$) plot (figure \ref{time_abs_timer_be}) we observe offsets and drift in the measurements, this time, caused by drift and misalignment between clocks. Note how messages from different modules arrive with different offset at the robot controller. Also we observe a drift, caused by the clocks of the modules running at a different rate than the robot controller's clock.\\ 


\subsubsection{Synchronizing system clocks with PTP}
\label{case3}

\begin{figure}[h!]
\centering
 \includegraphics[width=0.4\textwidth]{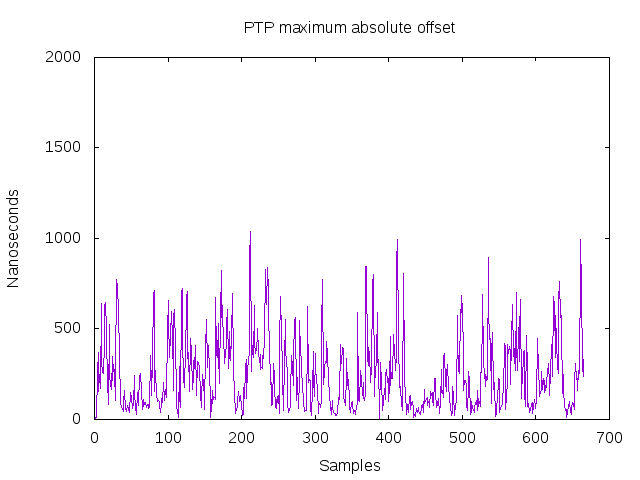}
\caption{\footnotesize Maximum absolute offset between clock synchronization for the system clock of rotary servo during 10 minutes. Each sample represents the maximum offset value for a second interval.}
\label{ptp_plot}
\end{figure}

In this case, we repeat the measurements with all the clocks synchronized. In figure \ref{ptp_plot} we can see how we get sub-microsecond accuracy observing the clock offsets between the PC and one of the module clocks for a 10 minutes long period. \\

\begin{figure}[h!]
\centering
 \includegraphics[width=0.4\textwidth]{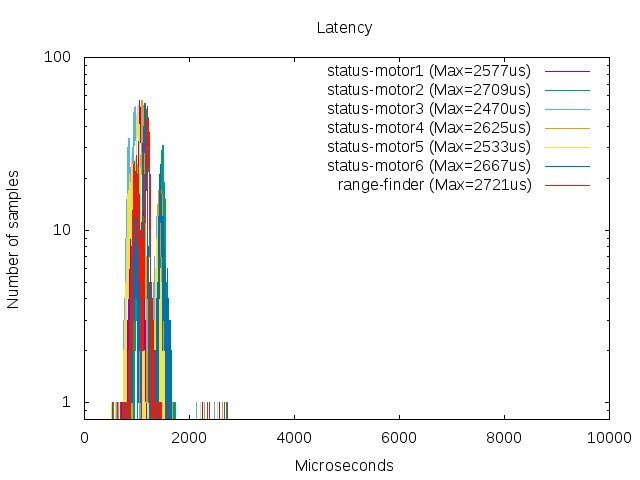}
\caption{\footnotesize Latency histogram for a 100 millisecond period with an absolute timer and synchronization.}
\label{hist_lat_abs_timer_ptp_be3}
\end{figure}

\begin{figure*}[h!]
  \begin{subfigure}[t]{.5\textwidth}
    \centering
    \includegraphics[width=0.8\linewidth]{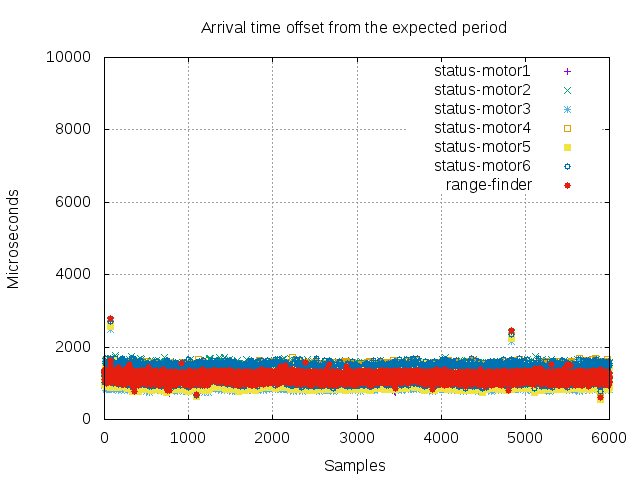}
    \caption{\footnotesize }
    \label{time_abs_timer_ptp_be3}
  \end{subfigure}
  \hfill
  \begin{subfigure}[t]{.5\textwidth}
    \centering
    \includegraphics[width=0.8\linewidth]{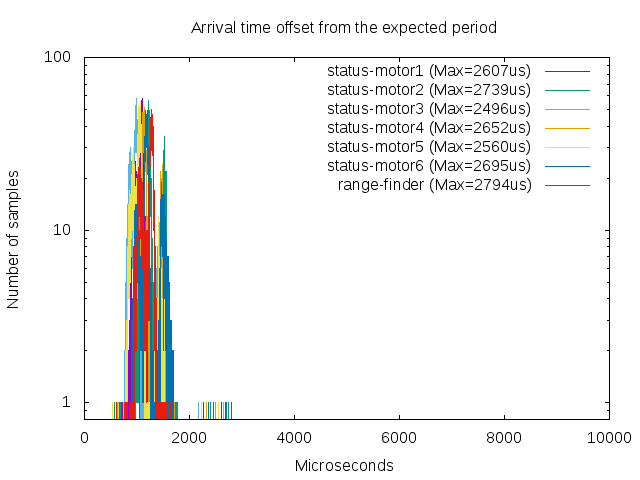}
    \caption{\footnotesize }
    \label{hist_rx_abs_timer_ptp_be3}
  \end{subfigure}
  \caption{\footnotesize Measurements for a 100 millisecond period and 10 minutes duration period. a) Arrival time offset ($\Delta t_{_{SUB}}$) timeplot with an absolute timer and synchronization. b) Arrival time offset ($\Delta t_{_{SUB}}$) histogram with an absolute timer and synchronization.}
\end{figure*}

In the arrival timestamp ($\Delta t_{_{SUB}}$) timeplot (figure \ref{time_abs_timer_ptp_be3}) we observe how all the messages arrive with the same offset and without any drift. The achieved synchronization between different topics is rather accurate. If a process was waiting for all the messages in order to merge them, for example using ROS \texttt{tf} filter, the synchronization delay to generate create the new data would be greatly reduced.\\



\begin{figure*}[h!]
  \begin{subfigure}[t]{.5\textwidth}
    \centering
    \includegraphics[width=0.8\linewidth]{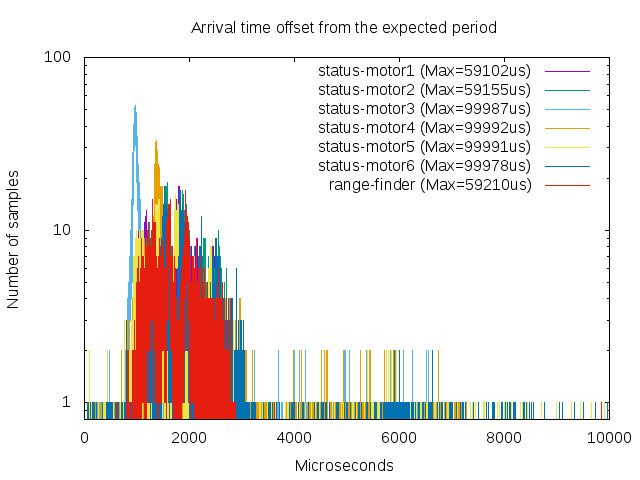}
    \caption{}
    \label{hist_rx_abs_timer_ptp_be_900}
  \end{subfigure}
  \hfill
  \begin{subfigure}[t]{.5\textwidth}
    \centering
    \includegraphics[width=0.8\linewidth]{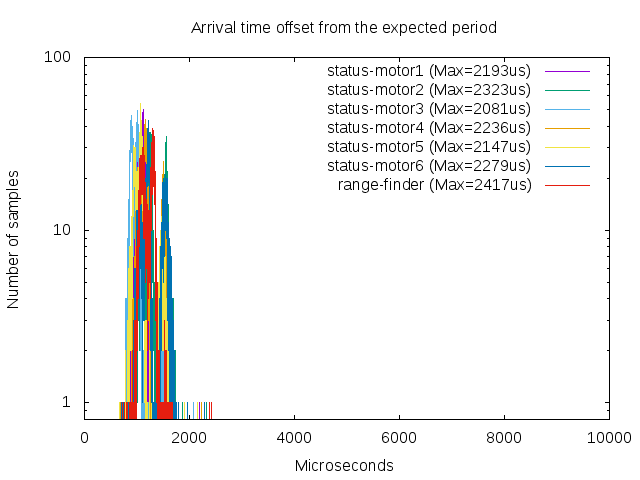}
    \caption{\footnotesize }
    \label{hist_rx_abs_timer_ptp_st_prio_900}
  \end{subfigure}
  \caption{Measurements for a 100 millisecond period and 10 minutes duration period with 90\% network congestion. a) Arrival time offset ($\Delta t_{_{SUB}}$) histogram. b) Arrival time offset ($\Delta t_{_{SUB}}$) histogram with traffic prioritization.}
\end{figure*}

\subsubsection{Ethernet link congested}

If the Ethernet link is congested packet loss and delay may occur. As shown in \cite{DBLP:journals/corr/abs-1804-07643} this effect is critical for traffic requiring low jitter transmission. For the current setup this delay becomes relevant for heavily congested links.\\

In this case we show the effects of the delay generated due to network congestion and how this is easily solved by using a Ethernet packet prioritization, also known as class of service (CoS). To generate congestion, we connect a bridged endpoint device between the robot arm and the controller, then we send contending traffic from a PC connected at the end of the robot arm. When we saturate the link to the 90\%, we observe that the communications are severely affected (figure \ref{hist_rx_abs_timer_ptp_be_900}). We repeat the test prioritizing the sensor and actuators traffic. As shown in figure \ref{hist_rx_abs_timer_ptp_st_prio_900}, this configuration solves the problem and the delays caused by the low priority traffic are almost eliminated.\\ 

Such congestion may not be a typical scenario for a 1 Gbps Mbps Ethernet link capacity, but in case of 100 Mbps links, it is not complicated to reach those saturation levels, for example, when using high resolution cameras.\\ 



\subsubsection{Using TSN Time-based packet transmission (Qbv)}

One of the TSN 802.1Qbv standard goals is to eliminate the delay caused by low priority traffic in switched Ethernet queues. This delay is negligible comparing with the jitter observed in this experiments. This implies that Qbv will not help reducing the jitter, because the jitter generated in the software layers is predominant. However, for this use case Qbv can also be used to schedule the transmission time for non-synchronized publishers. In many cases the ROS 2.0 publishers will not send their messages in a synchronized way based on the system clock. This might be the case for third party libraries and drivers, etc. In these cases we still can setup a Qbv schedule to control when traffic is transmitted in the endpoint egress queues.\\

\begin{figure}[h!]
\centering
 \includegraphics[width=0.4\textwidth]{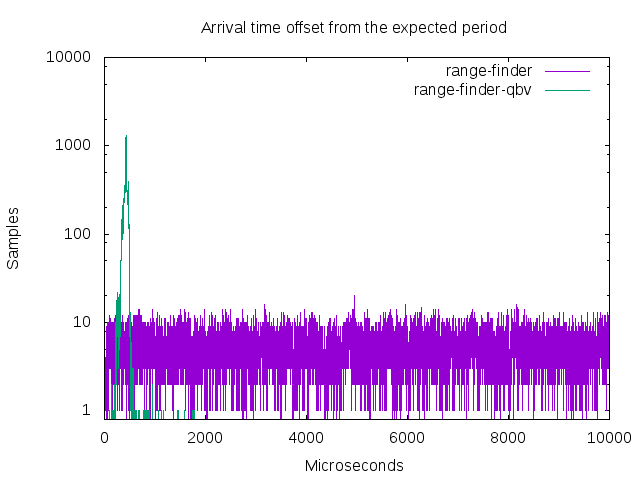}
\caption{\footnotesize Arrival time offset ($\Delta t_{_{SUB}}$) histogram for a 100 millisecond with and without TSN Time-based packet transmission (Qbv). }
\label{range_qbv}
\end{figure}

We repeat the experiments using the ROS 2.0 timers for the range finder publisher. The system clocks will be synchronized but the application will publish using the ROS 2.0 Timers, which are not aware of the synchronization. We configure a time window for this topic with a 10 millisecond cycle time and a zero microsecond offset from the start of the cycle. The publishing rate is configured to 10 milliseconds to match the Qbv cycle time \footnote{We used a 10 millisecond cycle time instead of 100 milliseconds because of maximum cycle time limitations with the Qbv implementation.}.\\

Comparing the arrival timestamp offset distribution with and without Qbv (figure \ref{range_qbv}), we observe how we can specify accurately when the range finder messages are transmitted, and hence when they arrive to the robot controller.\\

\section{Conclusion and future work}
\label{conclusions}

In this paper, our group researched the problem of time synchronization for modular robots. We describe how the lack of clock synchronization methods in robotic components is a critical problem that hampers the system integration effort and can affect the overall robot performance.


According to the results presented, we claim that the best approach for modular robotics is to build native synchronous robot modules which rely on standardized network-based synchronization protocols. We do so building on top of previous work \cite{DBLP:journals/corr/abs-1804-07643}, \cite{2018arXiv180810821G}, \cite{2018arXiv180407643G} integrated in the H-ROS\cite{hrim} infrastructure and prove how ROS-native robot components can turn into robot modules that provide synchronous responses and well defined communication latencies. In particular, we demonstrate how ROS 2.0-enabled hardware is able to obtain a) sub-microsecond clock synchronization accuracy (figure \ref{ptp_plot}), b) ROS 2.0 timespamping accuracies below 100 microseconds (figure \ref{hist_tx_abs_timer_be}) and c) end-to-end communication latencies between 1-2 milliseconds (figure \ref{hist_rx_abs_timer_ptp_be3}).\\

Overall, the results presented show that we achieved accurate modular and distributed synchronization using the setup described. We are able to timestamp ROS 2.0 messages with an accuracy under 100 microseconds. This is specially interesting for sensor data fusion based on timestamping, such as the ROS synchronization filters. We also showed how by using PTP, we are able to synchronize end-to-end ROS 2.0 communications with high accuracy, mainly limited by communications delays. As we showed in \cite{2018arXiv180407643G}, the communications overhead is mainly caused by the reception path of the Linux Network Stack and the lack of optimization of the ROS-DDS layers. Similar experiments with pub/sub communications for OPC UA over TSN are being conducted \cite{opcua_tsn} \cite{opcua_tsn_kalycito}, which show low latencies can be achieved when optimizations for time-sensitive traffic are made. In addition, we demonstrated how delays in the modules' Ethernet links can increase the synchronization jitter for ROS 2.0 messages in congested networks and showed that applying switched Ethernet QoS priorization solves the problem. Finally, we show that TSN Time Aware Shaper can be used in order to control the message transmission and achieve accurate reception times when ROS 2.0 nodes publish.\\







%

\bibliographystyle{IEEEtran}
\bibliography{references}


\onecolumn



%





\end{document}